\documentclass[conference]{IEEEtran}
\IEEEoverridecommandlockouts
\usepackage{amsthm}
\usepackage{amsfonts,amsmath}
\usepackage{algorithm}
\usepackage[noend]{algpseudocode}
\usepackage{bm}
\usepackage{booktabs}
\usepackage{soul}
\usepackage{multirow}
\usepackage{subfig}
\usepackage[flushleft]{threeparttable}
\usepackage{adjustbox}
\usepackage{fancyhdr}
\usepackage[dvipsnames]{xcolor}
\usepackage[flushleft]{threeparttable}
\usepackage{tikz}
\usepackage{outlines}
\usepackage{balance}
\usepackage{tcolorbox}
\usepackage{wrapfig}
\usepackage{bbm}
\usepackage{blindtext}
\usepackage{xcolor}
\usepackage{balance}
\usepackage{algorithm}
\usepackage{color}
\usepackage{listings}
\usepackage{color}
\usepackage{MnSymbol,wasysym}
\usepackage{marvosym}
\usepackage[switch]{lineno}
\usepackage{xcolor,colortbl}
\usepackage[labelsep=period]{caption}
\usepackage{cite}
\usepackage{MnSymbol,wasysym}
\usepackage{marvosym}
\usepackage{times}
\usepackage{caption}
\usepackage{textcase}

\usepackage{pifont}
\newcommand{\cmark}{\ding{51}}%
\newcommand{\xmark}{\ding{55}}%

\def\BibTeX{{\rm B\kern-.05em{\sc i\kern-.025em b}\kern-.08em
    T\kern-.1667em\lower.7ex\hbox{E}\kern-.125emX}}

\definecolor{Gray}{gray}{0.85}
\definecolor{LightCyan}{rgb}{0.88,1,1}

\usepackage{algorithm}
\usepackage{algpseudocode} 

\usepackage{twoopt}
\usepackage{tikz}
\usetikzlibrary{fit}
\begin{document}
\bstctlcite{IEEEexample:BSTcontrol}

\title{\Large \textbf{Graph Neural Network for Accurate and Low-complexity SAR ATR}
}

\author{
\IEEEauthorblockN{ Bingyi Zhang\IEEEauthorrefmark{1}, Sasindu Wijeratne\IEEEauthorrefmark{1}, Rajgopal Kannan\IEEEauthorrefmark{2}, Viktor Prasanna\IEEEauthorrefmark{1}, Carl Busart\IEEEauthorrefmark{2}}
\IEEEauthorblockA{
    \IEEEauthorrefmark{1}University of Southern California \IEEEauthorrefmark{2}DEVCOM US Army Research Lab\\
    e-mail: \IEEEauthorrefmark{1}\{bingyizh, kangaram, prasanna\}@usc.edu \IEEEauthorrefmark{2}\{rajgopal.kannan.civ, carl.e.busart.civ\}@army.mil}
}

\maketitle
\begin{abstract}
Synthetic Aperture Radar (SAR) Automatic Target Recognition (ATR) is the key technique for remote sensing image recognition. The state-of-the-art works exploit the deep convolutional neural networks (CNNs) for SAR ATR, leading to high computation costs. These deep CNN models are unsuitable to be deployed on resource-limited platforms.  In this work, we propose a graph neural network (GNN) model to achieve accurate and low-latency SAR ATR. We transform the input SAR image into the graph representation. The proposed GNN model consists of a stack of GNN layers that operates on the input graph to perform target classification. Unlike the state-of-the-art CNNs, which need heavy convolution operations, the proposed GNN model has low computation complexity and achieves comparable high accuracy. The GNN-based approach enables our proposed \emph{input pruning} strategy. By filtering out the irrelevant vertices in the input graph, we can reduce the computation complexity. Moreover, we propose the \emph{model pruning} strategy to sparsify the model weight matrices which further reduces the computation complexity. We evaluate the proposed GNN model on the MSTAR dataset and ship discrimination dataset. The evaluation results show that the proposed GNN model achieves 99.38\% and 99.7\% classification accuracy on the above two datasets, respectively. The proposed pruning strategies can prune 98.6\% input vertices and 97\% weight entries with negligible accuracy loss. Compared with the state-of-the-art CNNs, the proposed GNN model has only 1/3000 computation cost and 1/80 model size. 

\end{abstract}


\section{Introduction}
Synthetic aperture radar (SAR) is capable of high-resolution remote sensing and independent of weather conditions to observe the targets on the earth ground. 
SAR automatic target recognition (ATR) is the crucial technique to classify the target in the SAR images and has been used in many real-world applications, such as agriculture \cite{landuyt2018flood}\cite{zhan2021automated}, civilization \cite{li2021characterizing}\cite{zhang2020hyperli}, etc. SAR devices are typically mounted on moving platforms, such as aircraft, spacecraft, and small/micro satellites \cite{bardi2014integration, septanto2019simulation, yokota2013newly, akbar2016parallel, tanaka2018development}. These moving platforms usually have limited computation resources and power budgets (e.g., 80-180W \cite{tsamsakizoglou2017microsatellite}). The state-of-the-art works \cite{zhang2020convolutional, morgan2015deep, hu2018squeeze, pei2017sar, ying2020tai} develop complex convolutional neural networks (CNNs) for SAR ATR to achieve high classification accuracy. However, complex CNNs suffer from high computation costs and large memory footprints, making them unsuitable to be deployed on resource-limited platforms. For example, to achieve real-time image classification using CNNs, GPU is widely used. The power consumption of a state-of-the-art GPU device (e.g., NVIDIA RTX3090 has a power consumption of 450W) can exceed the power budget of the small/micro satellites.   

 \begin{figure}[ht]
    \centering
    \includegraphics[width=7cm]{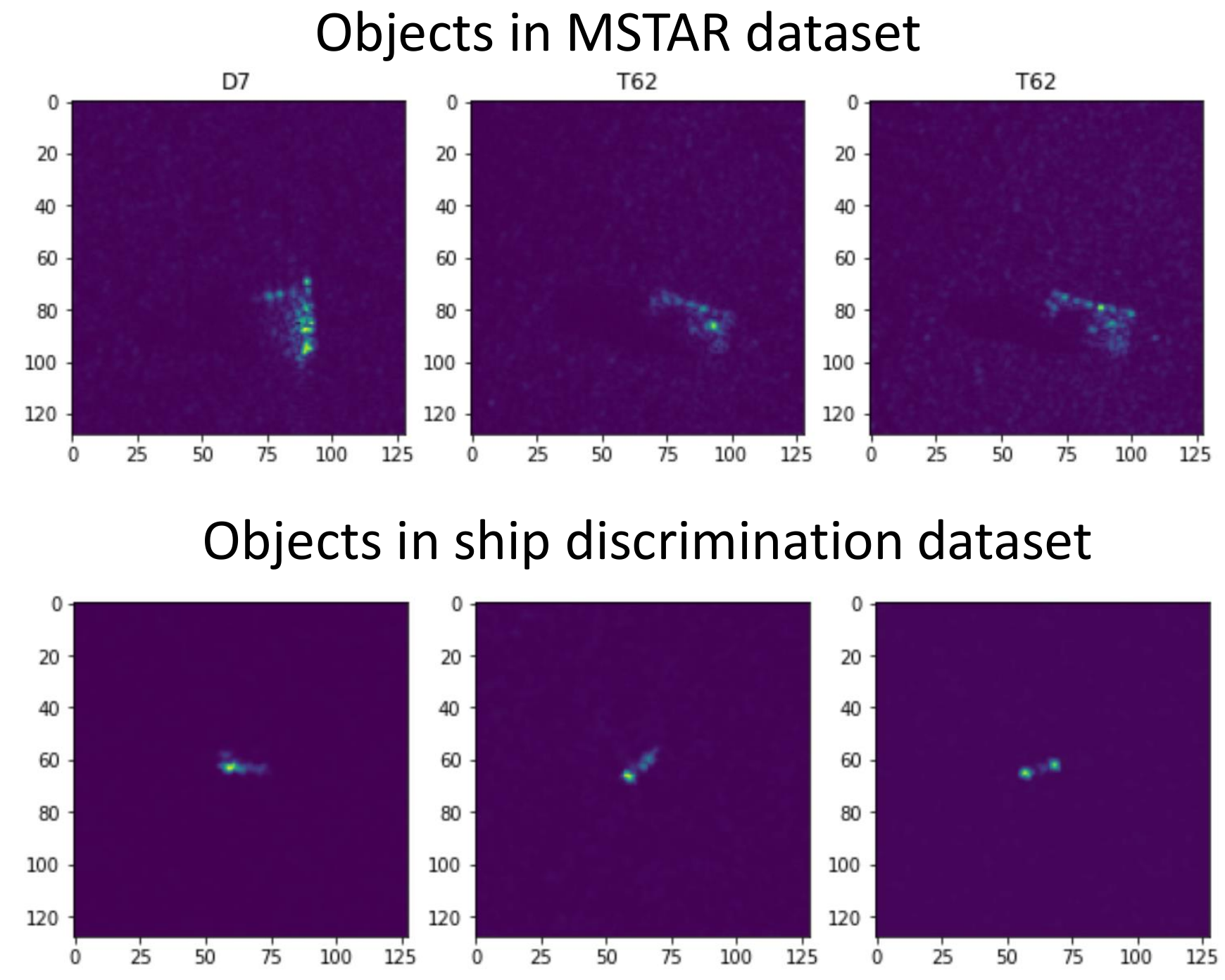}
    \captionsetup{justification=centering,font=rm}
    \caption{ The objects in the SAR images}
     \label{fig:sar-image}
\end{figure}

 \begin{figure*}[ht]
    \centering
    \includegraphics[width=18cm]{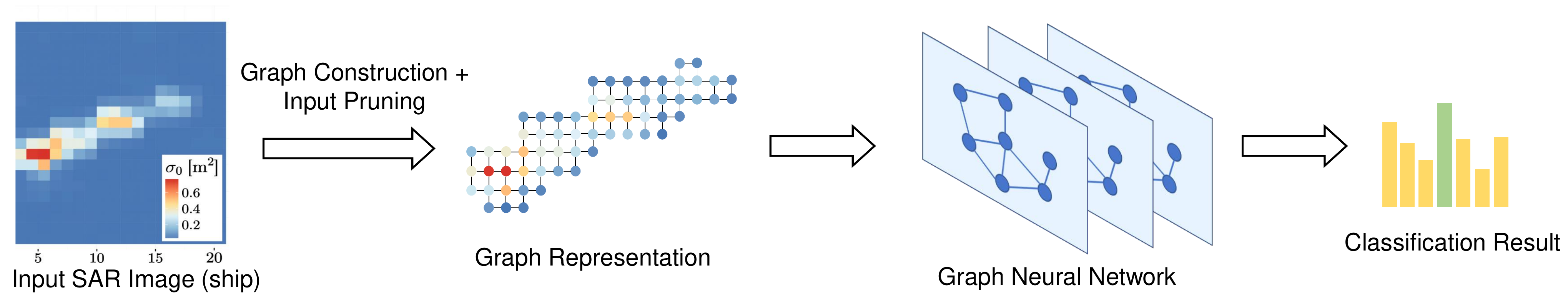}
    \captionsetup{justification=centering,font=rm}
    \caption{Overview of the proposed approach}
     \label{fig:overview}
\end{figure*}

We identify that CNNs have high computation costs due to (1) heavy convolution operations and (2) CNNs do not exploit the data sparsity in SAR images because CNNs need to use the whole image as input. As shown in Figure \ref{fig:sar-image}, an object in a SAR image usually has a small number of pixels, and most pixels are irrelevant for classification. Recently, Graph Neural Networks (GNNs) are proposed to operate on graph data structure and have been successfully applied to many graph classification tasks \cite{zhu2020convsppis, zhao2021identifying, qi2017pointnet++}, such as point cloud classification. \cite{xu2018powerful} has proven that GNN can classify a graph based on its graph structural information and vertex features. Motivated by that, we propose to use GNN for SAR ATR. First, we extract the image pixels of the target object. We use these pixels to build a graph by constructing the edge connections among the pixels. We exploit GNN to operate on the input graph for target classifying. The proposed GNN-based approach achieves significantly less computation cost and comparable accuracy compared with state-of-the-art CNNs. Moreover, we propose attention mechanisms, including vertex attention and feature attention, to improve the model's accuracy. Our main contributions are:
\begin{itemize}
    \item We propose a novel GNN model for SAR ATR with attention mechanisms, including vertex attention and feature attention, to achieve high accuracy with low computation complexity.
    \item We propose the input pruning strategy and the weight pruning strategy to further reduce the computation complexity with negligible accuracy loss. 
    \item We perform detailed ablation studies to evaluate (1) various connectivity for constructing the input graph, (2)  various types of GNN layers, (3) the effect of the attention mechanism, and (4) the impact of the proposed pruning strategies.
    \item We evaluate the proposed approach on MSTAR and ship discrimination datasets. The evaluation results show that the proposed GNN model achieves 99.38\% and 99.7\% classification accuracy on the above two datasets, respectively. Compared with the state-of-the-art CNNs, the proposed GNN model has only 1/3000 computation cost and 1/80 model size. 
\end{itemize}

The rest of the paper is organized as follows: 
Section \ref{Sec:Proposed-Model} presents the proposed GNN model for SAR ATR; 
Section \ref{sec::Pruning} describes the proposed pruning strategies for reducing computation complexity; Section \ref{sec:Evaluation} demonstrates the evaluation results. 
\section{Proposed Model}
\label{Sec:Proposed-Model}

Figure \ref{fig:overview} depicts the overview of the proposed approach. In Section \ref{subsec:GNN}, we introduce the basics of the graph neural network.  In Section \ref{subsec:graph-representation}, we cover the proposed graph representation for the SAR images. In Section \ref{subsec:model-architecture}, we introduce the proposed GNN model architecture.

\subsection{Graph Neural Network}
\label{subsec:GNN}

\begin{table}[]
\centering
\captionsetup{font=up}
\caption{NOTATIONS}
\begin{adjustbox}{max width=0.48\textwidth}
\begin{tabular}{cc|cc}
\toprule
 \textbf{{Notation}} & \textbf{{Description}}  & \textbf{{Notation}}  & \textbf{{Description}} \\
 \midrule
\midrule
{$  \mathcal{G}(\mathcal{V},\mathcal{E},\bm{X^{0}})$ }& {input graph}  &  $ v_{i}$ & {$i^{th}$ vertex} \\ \midrule
$ \mathcal{V}$ &  {set of vertices} &  $ e_{ij}$ & {edge from $ v_{i}$ to $  v_{j}$} \\ \midrule
$ \mathcal{E}$& {set of edges} &  $ L$&{number of GNN layers} \\ \midrule
$ \bm{h}_{i}^{l}$& feature vector of $ v_{i}$
at layer $l$    &  $ \mathcal{N}(i)$& {neighbors of $ v_{i}$} \\ 

 \bottomrule
\end{tabular}
\end{adjustbox}
\label{tab:notations}
\end{table}

We define GNN notations in Table \ref{tab:notations}. Graph Neural Networks  (GNNs) \cite{kipf2016semi, hamilton2017inductive, velivckovic2017graph} are proposed for representation learning on  graph $  \mathcal{G}(\mathcal{V},\mathcal{E},\bm{X}^{0})$. GNNs can learn from the structural information and vertex features and embed this information into low-dimension vector representation/graph embedding (For example, $\bm{h}^{L}_{i}$ is the embedding of vertex $v_{i}$). The vector representation can be used for many downstream tasks,  such as node classification \cite{hamilton2017inductive}\cite{kipf2016semi}, link prediction  \cite{zhang2018link}, graph classification \cite{ying2018hierarchical}, etc. As shown in Figure \ref{alg:GNN-computation-abstraction}, GNNs follow the message-passing paradigm that vertices recursively aggregate information from the neighbors.

\begin{figure}
\begin{algorithmic}[1]
 \renewcommand{\algorithmicrequire}{\textbf{Input:}}
\renewcommand{\algorithmicensure}{\textbf{Output:}}
 \Require Graph: $\mathcal{G}(\mathcal{V},\mathcal{E})$; vertex features: $\left\{\bm{h}^{0}_{1}, \bm{h}^{0}_{2}, ..., \bm{h}^{0}_{|\mathcal{V}|}\right\}$;
 \Ensure Output vertex features $\left\{\bm{h}^{L}_{1}, \bm{h}^{L}_{2}, ..., \bm{h}^{L}_{|\mathcal{V}|}\right\}$;
\For{$l=1...L$}
\For{each vertex $v \in \mathcal{V}$}
\State{$\bm{a}^l_{v} = {\text{Aggregate}(}\bm{h}_{u}^{l-1}: u\in \mathcal{N}(v))$}
\State{$\bm{z}_{v}^l = {\text{Update}(}\bm{a}_{v}^{l}, \bm{W}^{l} \textbf{)}$, $ \bm{h}_{v}^l = \sigma(\bm{z}_{v}^l )$}
\EndFor
\EndFor
\end{algorithmic}
\caption{GNN Computation Abstraction}
\label{alg:GNN-computation-abstraction}
\end{figure}

\subsection{Graph Representation}
\label{subsec:graph-representation}
\begin{figure}[h]
    \centering
    \includegraphics[width=4.5cm]{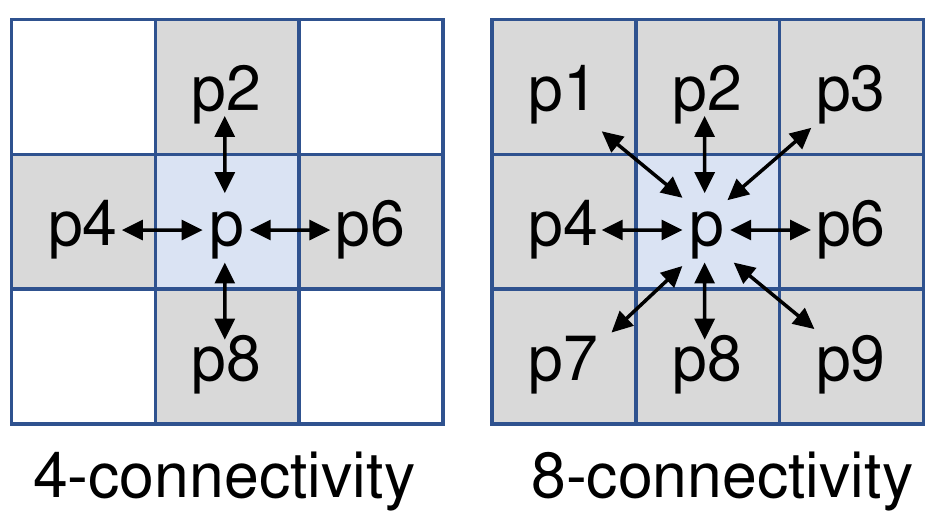}
    \captionsetup{justification=centering,font=rm}
    \caption{Two types of connectivity for constructing input graph}
     \label{fig:connectivity}
\end{figure}

We transform the input SAR image into a graph representation $  \mathcal{G}(\mathcal{V},\mathcal{E},\bm{X^{0}})$, where each pixel in the SAR image is mapped to a vertex $v\in \mathcal{V}$ in the graph. The SAR signal value of the pixel becomes the feature of the vertex.
Each pixel is connected to its neighbors as the edge connections $\mathcal{E}$. As shown in Figure \ref{fig:connectivity}, we propose the following two ways of connecting a pixel to its neighbors and evaluate them in the experiments:
\begin{itemize}
    \item \textbf{4-connectivity}: Each pixel is connected to the four neighbors: up ($p2$), down ($p8$), left ($p4$), and right ($p6$).
    \item \textbf{8-connectivity}: Each pixel is connected to the eight neighbors: $p1$, $p2$, $p3$, $p4$, $p6$, $p7$, $p8$, $p9$.
\end{itemize}

\subsection{Model Architecture}
\label{subsec:model-architecture}

 \begin{figure}[h]
    \centering
    \includegraphics[width=8.5cm]{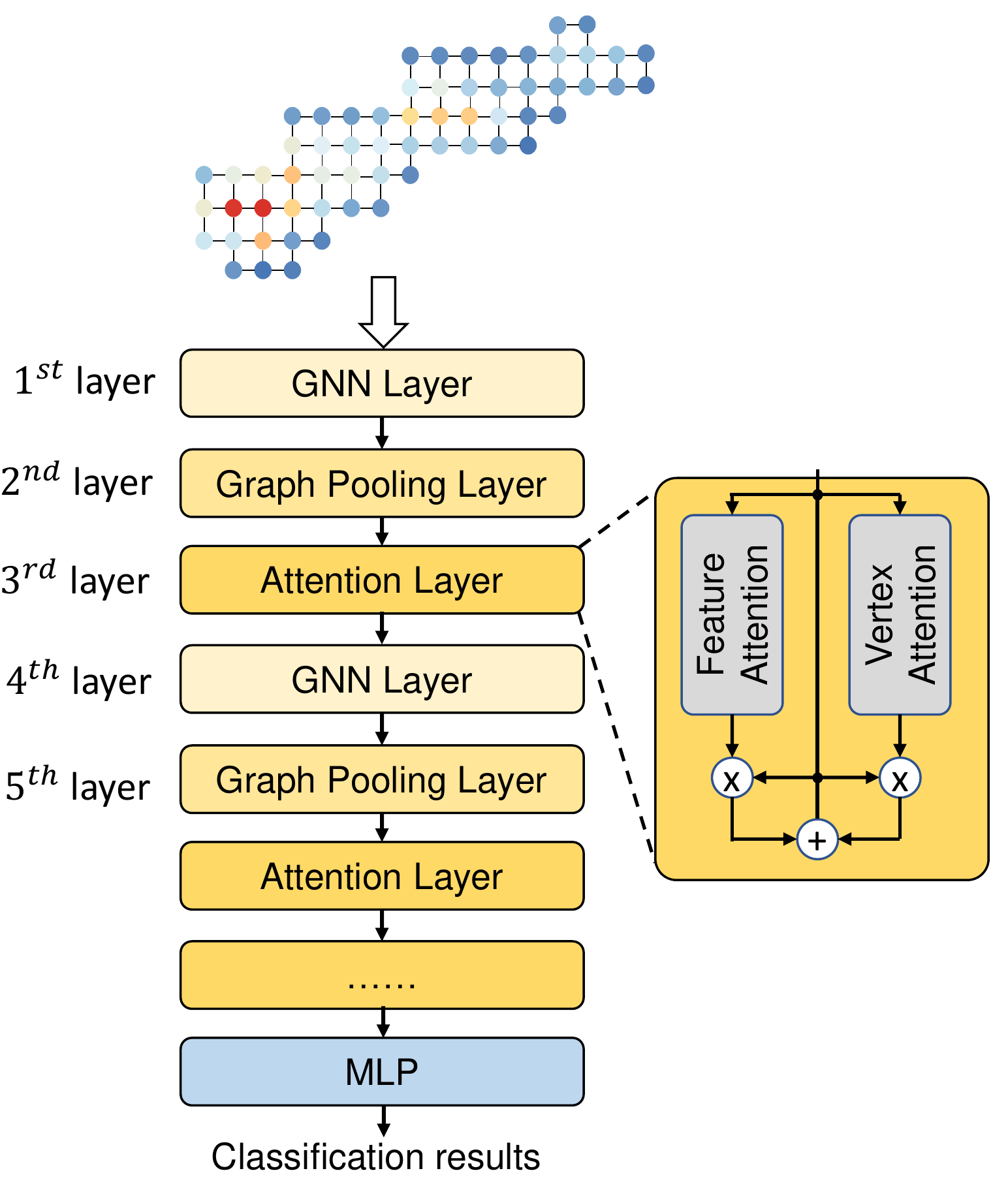}
    \captionsetup{justification=centering,font=rm}
    \caption{Diagram of model architecture}
     \label{fig:model-architecture}
\end{figure}

The proposed model architecture is shown in Figure \ref{fig:model-architecture}, which consists of a stack of layers, including Graph Neural Network layers, graph pooling layers, and attention layers. The final Multi-layer Perceptron (MLP) generates the classification result. For simplicity, $v_{i,j}$ denotes the vertex/pixel that locates at $i^{\text{th}}$ row and $j^{\text{th}}$ column in original SAR image. The input to layer $l$ $(1 \leqslant l \leqslant L)$ is the vertex feature vectors $\{\bm{h}_{i,j}^{l-1}: v_{i,j}\in \mathcal{V}_{l-1}\}$ and edges $\{e: e \in \mathcal{E}_{l-1} \}$ that defines the connectivity of the vertices in $\mathcal{V}_{l-1}$. The output of layer $l$ is the vertex feature vectors  $\{\bm{h}_{i,j}^{l}: v_{i,j}\in \mathcal{V}_{l}\}$. 

\vspace{0.1cm}
\noindent \textbf{Graph neural network (GNN) layer}: A GNN layer follows the \emph{Aggregate-Update} paradigm as shown in Algorithm \ref{alg:GNN-computation-abstraction}. Using the Aggregate() function, each vertex aggregates the feature vectors from the neighbors (line 3 of Algorithm \ref{alg:GNN-computation-abstraction}). Then, each feature vector is updated by the Update() function to generate the updated feature vector (line 4 of Algorithm \ref{alg:GNN-computation-abstraction}). There are some representative Graph Neural Network layers, such as GCN \cite{kipf2016semi}, GraphSAGE \cite{hamilton2017inductive}, GIN \cite{xu2018powerful}, and SGC \cite{wu2019simplifying}.

\vspace{0.1cm}
\noindent \textbf{Graph pooling layer}: It downscales the input graph $\mathcal{V}_{l-1}$ into a smaller output graph $\mathcal{V}_{l}$. The pooling operaton is similar to the pooling in the 2-D images:
\begin{equation}
\label{eq:pooling-op}
\bm{h}_{i,j}^{l} = \max(\bm{h}_{2i,2j}^{l -1}, \bm{h}_{2i + 1,2j}^{l -1}, \bm{h}_{2i,2j+1}^{l -1}, \bm{h}_{2i+1,2j+1}^{l -1})
\end{equation}
where $v_{i,j}^{l} \in \mathcal{V}_{l}$, and $v_{2i,2j}^{l -1}, v_{2i + 1,2j}^{l -1}, v_{2i,2j+1}^{l -1}, v_{2i+1,2j+1}^{l -1} \in \mathcal{V}_{l-1}$.

\vspace{0.1cm}
\noindent \textbf{Attention layer}: We exploit the attention mechanism to improve the accuracy. The attention layer consists of \emph{feature attention} that calculates the attention scores for each vertex feature, and \emph{vertex attention} that calculates the attention scores for each vertex. The feature attention is calculated by: 
\begin{equation}
    \begin{split}
        \bm{F}_{\text{fa}} = \text{sigmoid} (\text{mean}(\{\bm{h}_{i,j}:v_{i,j}\in \mathcal{V}\}) \bm{W}_{\text{fa}}^{\text{mean}} +  \\ \text{sum}(\{\bm{h}_{i,j}:v_{i,j}\in \mathcal{V}\}) \bm{W}_{\text{fa}}^{\text{sum}}) 
    \end{split}
\end{equation}
where $\bm{h}_{i,j}, \bm{F}_{\text{fa}}\in \mathbb{R}^{c}, ~\bm{W}_{\text{fa}}^{\text{mean}}, \bm{W}_{\text{fa}}^{\text{sum}}\in \mathbb{R}^{c\times c}$, and $c$ denotes the length of feature vector. $\text{fa}[i]$ is the attention score for $i^{\text{th}}$ feature. The vertex attention score is calculated using a GNN layer:
\begin{equation}
    \{\alpha_{i,j}:v_{i,j}\in \mathcal{V}_{l}\} = \text{sigmoid} (\text{GNNL}(\{\bm{h}_{i,j}:v_{i,j}\in \mathcal{V}_{l-1}\})),
\end{equation}
Where $\alpha_{i,j}$ is the attention score for vertex $v_{i,j}$. Then, the output of the attention layer is calculated by:
\begin{equation}
    \{\bm{h}_{i,j}^{\text{out}}: \bm{h}_{i,j}^{\text{out}} = (1+\alpha_{i,j})\bm{h}_{i,j}^{\text{in}} + \bm{h}_{i,j}^{\text{in}}\otimes\bm{F}_{\text{fa}}   \}
\end{equation}
where $\otimes$ is element-wise multiplication.

\vspace{0.1cm}
\noindent \textbf{Multi-layer Perceptron (MLP)}: After a sequence of layers, all the feature vectors are flattened into a single vector, which is sent to the MLP for classification. MLP has a stack of fully connected (FC) layers.

\section{Pruning}
\label{sec::Pruning}

This section covers the proposed pruning techniques, including, input pruning (Section \ref{subsec::input-pruning}), and weight pruning (Section \ref{subsec:weight-pruning}).

\subsection{Input Pruning}
\label{subsec::input-pruning}
The key benefit of using GNN is that GNN is flexible in accepting any graph structure as the input. Thereby, we are able to exploit input pruning to reduce the computation complexity. Theoretically, in a SAR image (See Figure \ref{fig:sar-image}), the pixels not in the target do not affect the classification results. As studied in \cite{zhu2020target}, by properly setting up a constant threshold $I_{v}$, we can filter out most irrelevant pixels since the pixels that do not belong to the target usually have negligible SAR signal magnitude. After constructing the input graph from the SAR image, we prune the vertices that have a magnitude smaller than $I_{v}$. The magnitude of a vertex is calculated by $\sqrt{x^{2}_{1} + x^{2}_{2} + ... + x^{2}_{np}}$ where $np$ denotes the number of polarization of the SAR signal. For example, a quad-polarization system has four kinds of polarization -- horizontal-horizontal (HH), vertical-vertical (VV), horizontal-vertical (HV), and vertical-horizontal (VH). After pruning the vertices, all the edges connected to the pruned vertices are also pruned. Due to the input pruning, the graph pooling operation (Equation \ref{eq:pooling-op}) is slightly modified:
\begin{equation}
\begin{split}
    \bm{h}_{i,j}^{l} = \max(\mathbbm{1}_{2i,2j}^{l -1} \cdot \bm{h}_{2i,2j}^{l -1}, \mathbbm{1}_{2i + 1,2j}^{l -1} \cdot  \bm{h}_{2i + 1,2j}^{l -1}, \\
    \mathbbm{1}_{2i,2j+1}^{l -1} \cdot\bm{h}_{2i,2j+1}^{l -1}, \mathbbm{1}_{2i+1,2j+1}^{l -1} \cdot \bm{h}_{2i+1,2j+1}^{l -1}),    
\end{split}
\end{equation}
where $\mathbbm{1}_{i,j} \in \{0,1\}$ is the indicator that indicates the existence of vertex $v_{i,j}$. After input pruning, we can skip the computation for the pruned vertices, which greatly reduces the total computation complexity.

\subsection{Weight Pruning}
\label{subsec:weight-pruning}

As analyzed in \cite{rahman2022triple, zhou2021accelerating}, the weight matrices in GNNs have redundancy, and some weight entries can be pruned without affecting the classification accuracy. Therefore, to reduce the total computation complexity, we perform weight pruning by training the model using lasso regression \cite{tibshirani1996regression}. We add the L1 penalty to the loss function:
\begin{equation}
\label{eq:loss}
    \text{loss} = l(y, y') + \lambda \sum_{w}^{W}|w|
\end{equation}
where $l(y, y')$ is the classification loss, and $\lambda \sum_{w}^{W}|w|$ is the L1 penalty term parameterized by $\lambda$. The L1 penalty leads to weight shrinkage during training. Thereby, some model weights become zeros and can be eliminated from the model. After training, we set a threshold $I_{w}$, and the model weights with absolute values smaller than $I_{w}$ are pruned.

\section{Evaluation}
\label{sec:Evaluation}

We evaluate our approach on two widely used datasets:
\begin{itemize}
    \item \textbf{MSTAR}: The setting of the MSTAR dataset follows the state-of-the-art work \cite{zhang2020convolutional}\cite{pei2017sar}\cite{ying2020tai}\cite{morgan2015deep}. MSTAR contains the SAR images of ten classes of ground vehicles, with 2747 images in the training set and 2427 images in the testing set.
    \item \textbf{Ship discrimination \cite{schwegmann2016very}:} For the ship discrimination dataset, we follow the setting in \cite{rostami2019deep}, which is a binary classification task that identifies if a given SAR image has a ship or not. The dataset contains 1596 positive image samples and 1596 negative image samples.
\end{itemize}

\subsection{Evaluation on MSTAR Dataset}

\subsubsection{Experimental Setting} For the MSTAR dataset, we use the following setting. The proposed model consists of  12 layers. We develop the proposed model using Pytorch Geometric. We use the cross-entropy loss as the classification loss (Equation \ref{eq:loss}). We train the model using the Adam optimization algorithm. The training batch size is set as $20$, and the initial learning rate is $0.02$. $\lambda$ (for lasso regression) is set as $0.002$. The L2 weight decay is set as $0.08$. We train the model for $150$ epochs, and the learning rate is multiplied by $0.5$ for every $10$ epoch. We use the $8$-connectivity to build the input graph. We evaluate the three widely used GNN layers in the proposed model -- GCN layer \cite{kipf2016semi}, GraphSAGE layer \cite{hamilton2017inductive}, and GAT \cite{velivckovic2017graph}. We train the proposed model using one NVIDIA RTX A6000 GPU.

\vspace{0.1cm}
\noindent \textbf{Performance metrics}: We evaluate the proposed approach using the following metrics: \emph{classification accuracy}, \emph{computation complexity}, and \emph{number of parameters}.

\begin{table}[H]
\centering
\caption{THE ACCURACY ON MSTAR DATASET}
\begin{adjustbox}{max width=0.45\textwidth}
\begin{tabular}{ccccc}
\toprule
\begin{tabular}[|c|]{@{}c@{}}  \textbf{GNN Layer} \\ \textbf{Type} \end{tabular}  & \textbf{Connectivity} & \begin{tabular}[|c|]{@{}c@{}}  \textbf{Training} \\ \textbf{Accuracy} \end{tabular}  &  \begin{tabular}[|c|]{@{}c@{}}  \textbf{Testing} \\ \textbf{Accuracy} \end{tabular}  &  \begin{tabular}[|c|]{@{}c@{}}  \textbf{Training} \\ \textbf{Time} \end{tabular} \\
\midrule
\midrule
\multirow{2}{*}{GCN}  & $4$ & $99.16\%$ & $90.06\%$ & $3.0$ hours \\
                      & $8$ & $95.44\%$ & $83.82\%$ & $4.0$ hours \\
\midrule
\multirow{2}{*}{GAT}  & $4$ & $99.53\%$ &  $92.21\%$ & $1.8$ hours \\
                      & $8$ & $82.71\%$ &  $71.33\%$ & $1.9$ hours \\
\midrule
\multirow{2}{*}{GraphSAGE} & $4$ & $100.00\%$  &  $97.81\%$ &  $52$ min\\
                      & $8$ & \textbf{$100.00\%$}  &  \textbf{$99.38\%$} & $55$ min\\
\bottomrule
\end{tabular}
\end{adjustbox}
\label{tab:accuracy-results}
\end{table}

\subsubsection{Classification Accuracy}  The accuracy of the proposed model (under various GNN layer types and connectivity) is shown in Table \ref{tab:accuracy-results}. We observe that using the GraphSAGE layer as the GNN layer leads to the highest training/testing accuracy. Using the GraphSAGE layer also leads to the lowest training time. For the GraphSAGE layer, using 8-connectivity to build the input graph can result in higher accuracy but slightly higher training time than 4-connectivity. Table \ref{tab:comparsion-state-of-the-art} shows that the proposed GNN model achieves higher accuracy compared with the state-of-the-art CNNs \cite{zhang2020convolutional, pei2017sar, ying2020tai, morgan2015deep} with negligible computation complexity for inference.

\begin{table}[!ht]
\centering
\caption{COMPARISON WITH THE STATE-OF-THE-ART CNNS ON MSTAR DATASET}
\begin{adjustbox}{max width=0.48\textwidth}
\begin{tabular}{ccccc}
\toprule
 & Type & Accuracy & \# of FLOPs & \# of Para.\\ \midrule \midrule
 \cite{zhang2020convolutional}&CNN&92.3\%&$\frac{1}{12}\times$&$0.5 \times 10^{6}$\\ \midrule
\cite{pei2017sar} &CNN&97.97\%&$\frac{1}{10}\times$&$0.65 \times 10^{6}$  \\  \midrule
\cite{ying2020tai}&CNN&98.52\%&$\frac{1}{3}\times$&$2.1 \times 10^{6}$   \\  \midrule
\cite{morgan2015deep}&CNN&99.3\%&  \begin{tabular}[|c|]{@{}c@{}} $1\times$ \\ (6.94 GFLOPs) \end{tabular} &$2.5 \times 10^{6}$  \\ \midrule
 \begin{tabular}[|c|]{@{}c@{}}This work [after pruning] \\ (GraphSAGE layer, \\ 8-connectivity) \end{tabular}  &GNN&99.1\%&$\frac{1}{3000}\times$&$0.03\times 10^{6}$  \\ \bottomrule
\end{tabular}
\end{adjustbox}
\label{tab:comparsion-state-of-the-art}
\end{table}

\begin{table}[!ht]
\centering
\caption{THE IMPACT OF THE ATTENTION MECHANISM (USING GRAPHSAGE LAYER AND 8-CONNECTIVITY)}
\begin{tabular}{ccccc}
\toprule
\begin{tabular}[|c|]{@{}c@{}}  \textbf{Vertex} \\ \textbf{Attention} \end{tabular}                   & \begin{tabular}[|c|]{@{}c@{}}  \textbf{Feature} \\ \textbf{Attention} \end{tabular}  & \begin{tabular}[|c|]{@{}c@{}}  \textbf{Training} \\ \textbf{Accuracy} \end{tabular}  & \begin{tabular}[|c|]{@{}c@{}}  \textbf{Testing} \\ \textbf{Accuracy} \end{tabular}  &  \begin{tabular}[|c|]{@{}c@{}}  \textbf{Training} \\ \textbf{Time} \end{tabular}  \\ 
\midrule 
\midrule
 \xmark  &  \xmark & $99.67\%$ & $93.77\%$ & 31 min \\
 \xmark  &  \cmark & $100.0\%$ & $98.51\%$ & 40 min \\
 \cmark  &  \xmark & $100.0\%$ & $99.26\%$ & 41 min \\
 \cmark  &  \cmark & $100.0\%$ & $99.38\%$ & 55 min \\
\bottomrule
\end{tabular}
\label{tab:impact-attention}
\end{table}

\subsubsection{Ablation Study} We perform an ablation study to evaluate the impact of the attention mechanism (using GraphSAGE layer and 8-connectivity). The result is shown in Table \ref{tab:impact-attention}. Without vertex and feature attention, the model achieves only $93.77\%$ accuracy. With only vertex attention, the model achieves $99.26\%$ accuracy. With only feature attention, the model achieves $98.51\%$ accuracy. With both vertex and feature attention, the model achieves $99.38\%$ accuracy. The evaluation result demonstrates that the attention mechanism can improve classification accuracy without significantly increasing computation complexity.

\subsubsection{Evaluation on the Pruning Strategy} 

 \begin{figure}[h]
    \centering
    \includegraphics[width=4.2cm]{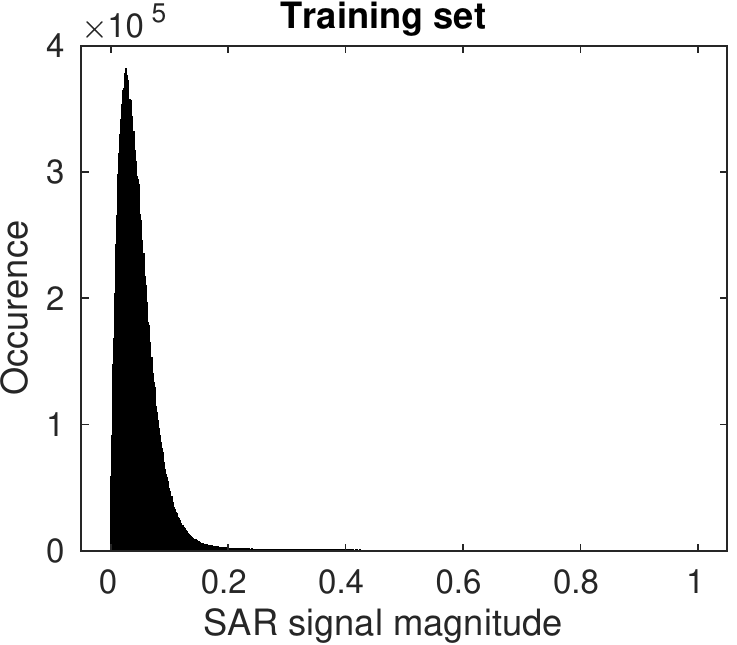}
    \includegraphics[width=4.2cm]{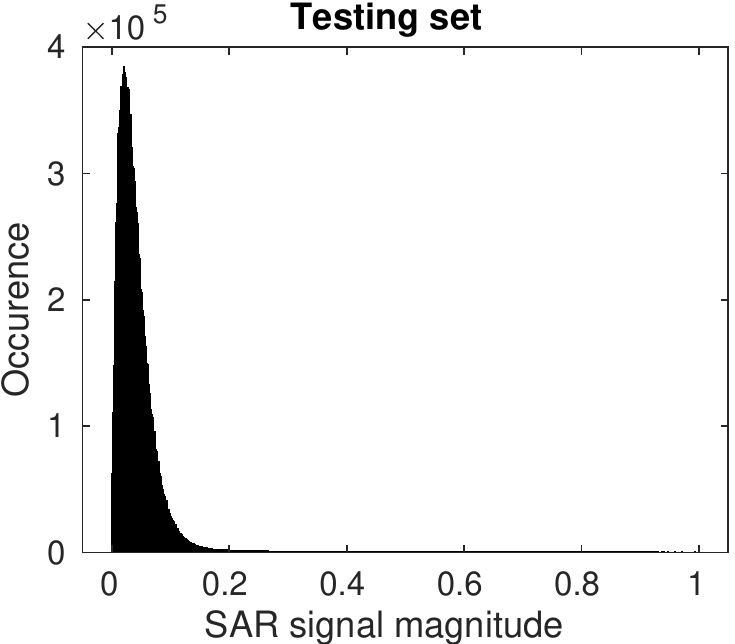}
    \caption{The distribution of the SAR signal magnitude in the training/testing set of MASTAR}
     \label{fig:histogram}
\end{figure}

 \begin{figure*}[h]
    \centering
    \includegraphics[width=18cm]{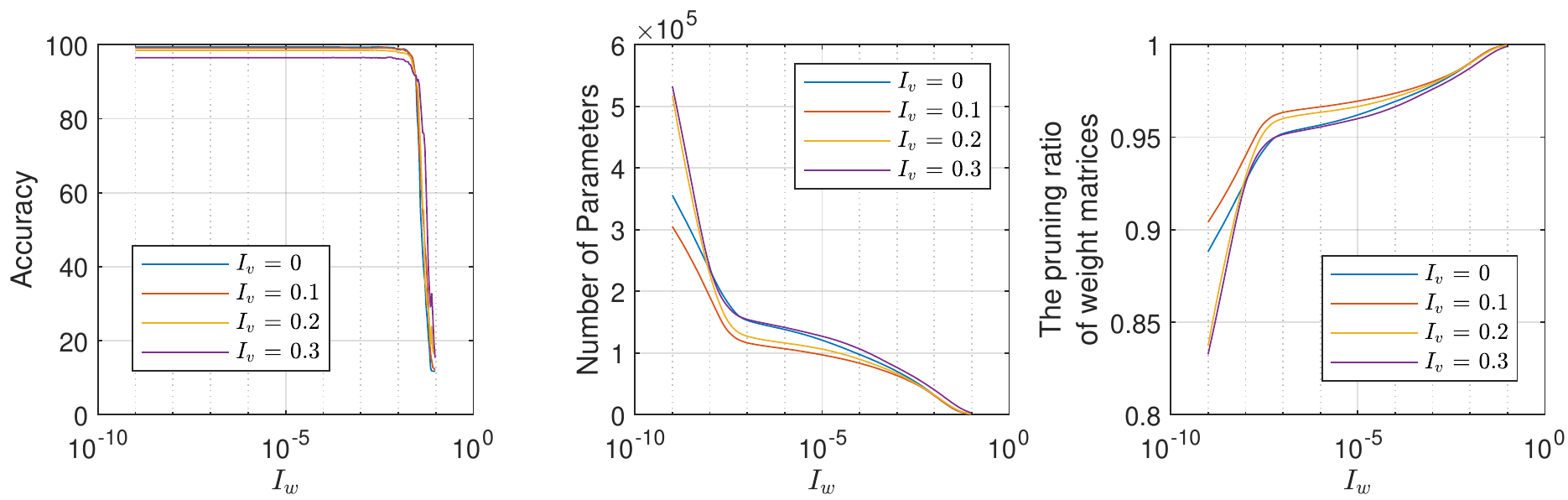}
    \captionsetup{justification=centering,font=rm}
    \caption{Evaluation of proposed pruning strategy}
     \label{fig:pruning-eva}
\end{figure*}

We evaluate the proposed input pruning and weight pruning strategies. We use GraphSAGE layer and 8-connectivity as the setting of the model.

\vspace{0.1cm}
\noindent \textbf{Input Pruning}: Figure \ref{fig:histogram} shows the data distribution of the SAR signal magnitude of the image pixels in the training/testing set. The SAR signal magnitude ranges from $0$ to $16$. Since most pixels have a magnitude between $0$-$1$, Figure \ref{fig:histogram} only shows the range $0$-$1$. For experiment, we set the pruning threshold $I_{v}$ (See Section \ref{subsec::input-pruning}) to be $0$, $0.1$, $0.2$, $0.3$ respectively. The image pixels that have a magnitude small than $I_{v}$ are pruned.

\vspace{0.1cm}
\noindent \textbf{Weight Pruning}:  The weights in weight matrices can be either negative or positive. We set the threshold $I_{w}$ for weight pruning (See Section \ref{subsec:weight-pruning}). The weights that have an absolute value that is smaller than $I_{w}$ are pruned. In the experiment, we set $I_{w}$ to be between $1\times 10^{9}$ and $1\times 10^{1}$.

\vspace{0.1cm}
The evaluation results for the pruning strategy are shown in Figure \ref{fig:pruning-eva}. We have the following observations:
\begin{itemize}
    \item Without weight pruning, when $I_{v} = 0.1$, $93.4\%$ input vertices/pixels are pruned, the accuracy is dropped to $99.1\%$; when $I_{v} = 0.2$, $98.6\%$ input vertices/pixels are pruned, the accuracy is dropped to $98.5\%$; when $I_{v} = 0.3$, $99.1\%$ input vertices/pixels are pruned, the accuracy is dropped to $96.5\%$.
    \item When weight pruning threshold $I_{w} < 10^{7}$, the accuracy does not change w.r.t. to $I_{w}$. When $I_{w} = 10^{7}$, more than $95\%$ weights are pruned. Therefore, most entries in the weight matrices are redundant. 
\end{itemize}

Therefore, by setting proper threshold $I_{v}$, $I_{w}$ for input pruning and weight pruning, most input pixels and weights can be pruned without significantly dropping the accuracy. Figure \ref{fig:pruning-eva} shows the evaluation results for the pruning strategy, $97\%$ weight entries are pruned,  and the accuracy is $99.1\%$. By skipping the computation for the pruned vertices and weights, we can dramatically reduce the total computation complexity.


\subsubsection{Experimental Setting} For ship discrimination dataset, we follow the setting of \cite{rostami2019deep} to conduct experiment for few-shot learning.  Since the ship discrimination is a binary class task, the few-shot learning task can be formed as a 2-way-$K$-shot-classification problem, where $K=\{1,2,..,10\}$ denotes the number of labeled training images for each class.  We train the model using the Adam optimization algorithm. The training batch size is set as $\frac{K}{2}$, and the learning rate is set as $0.001*K$. The L2 weight decay is set as $0.08$.

\subsubsection{Classification Accuracy} 
\begin{figure}[h]
    \centering
    \includegraphics[width=7cm]{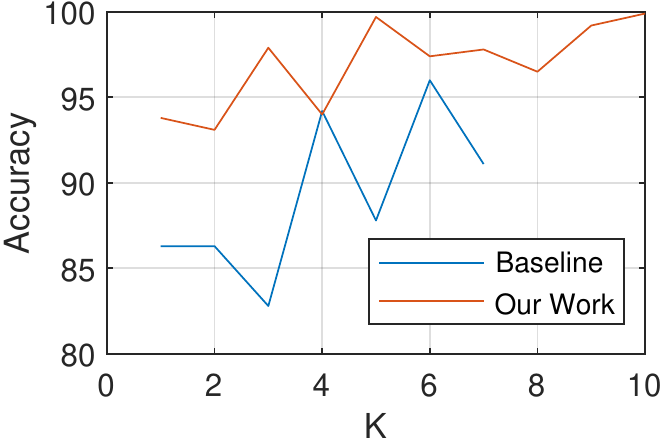}
    \captionsetup{justification=centering,font=rm}
    \caption{The accuracy on the ship discrimination dataset}
     \label{fig:ShipAccuracy}
\end{figure}

As shown in Figure \ref{fig:ShipAccuracy}, we compare our accuracy with \cite{rostami2019deep} (baseline) for the few-shot learning on the ship discrimination dataset. Note that the baseline \cite{rostami2019deep} uses a convolutional neural network (CNN), and the authors pretrained their CNN using the ship discrimination dataset on the Electro-Optical (EO) domain. We do not pretrain our network on any dataset. For various $K$, the proposed model outperforms the baseline \cite{rostami2019deep}, which is a pretrained deep CNN model.

\begin{table}[!ht]
\centering
\caption{COMPARISON OF ACCURACY (\%)}
\begin{tabular}{lllllllllll}
\toprule
 $K$ & 1 & 2 & 3 & 4 & 5 & 6 & 7 \\
 \midrule
 \midrule
 Baseline \cite{rostami2019deep} & 86.3 & 86.3 & 82.8  & 94.2 & 87.8 & 96.0 & 91.1  \\
 Our work & 93.8 & 93.1 & 97.9 & 94.0 & 99.7 & 97.4 &  97.8 \\
 \bottomrule
\end{tabular}
\end{table}
\section{Conclusion and Future Work}
In this paper, we proposed a novel GNN-based approach for SAR automatic target recognition. The proposed approach uses the GNN layer as the backbone and uses the attention mechanism to improve classification accuracy. We proposed pruning strategies, including input pruning and weight pruning, to reduce the computation complexity. The evaluation results on the MSTAR and ship discrimination datasets show that the proposed model outperforms the state-of-the-art CNNs in classification accuracy and computation complexity. In \cite{10035150}, we designed a hardware accelerator for the proposed GNN model. In the future, we plan to extend the proposed GNN model to more SAR-related tasks, such as object detection.

\section*{Acknowledgment}
This work is supported by the National Science Foundation (NSF) under grants CCF-1919289 and OAC-2209563, and  the  DEVCOM  Army Research Lab (ARL) under grant W911NF2220159. 

\bibliographystyle{IEEEtran}
\bibliography{main_full}

\end{document}